\def\lmjcorespemail{...@...}
\def\lmjrunauthor{...}
\def\lmjruntitle{...}

\def\lmjmsc{}
\def\lmjkeywords{Fine-tuning large language model, LoRA (Low-Rank Adaptation), Alignment}

\documentclass{lmj}
\include{lmj.def}
\LMJCLS%
\usepackage{multirow}
\usepackage{booktabs}
\usepackage{multirow}
\usepackage{caption} 
\begin{document}

\setlength{\parskip}{0.5em} 
\setlength{\parindent}{1em} 
\setlength{\topmargin}{-1cm} 
\setlength{\textheight}{23cm} 
\setlength{\headsep}{0.5cm} 

\setlength{\abovedisplayskip}{8pt}
\setlength{\belowdisplayskip}{8pt}
\setlength{\abovedisplayshortskip}{6pt}
\setlength{\belowdisplayshortskip}{6pt}
\setlength{\jot}{3pt} 

\renewcommand{\normalsize}{\fontsize{12pt}{14pt}\selectfont}
\renewcommand{\small}{\fontsize{11pt}{13pt}\selectfont}
\renewcommand{\footnotesize}{\fontsize{10pt}{12pt}\selectfont}
\renewcommand{\large}{\fontsize{14pt}{17pt}\selectfont}
\renewcommand{\Large}{\fontsize{16pt}{19pt}\selectfont}
\renewcommand{\LARGE}{\fontsize{18pt}{22pt}\selectfont}

\setlength{\footnotesep}{3pt} 

\renewcommand{\normalsize}{\fontsize{12pt}{14pt}\selectfont}

\pagestyle{plain} 
\bibliographystyle{plainlmj}
 \title{Slimming Down LLMs Without Losing Their Minds}
 \author{Qingda (Michael) Mai}

  \institution{University of Waterloo\\Department of Management Science and Engineering}

 \email{michael.mai@uwaterloo.ca}
%
\LMJarticle
 \begin{abstract}

This paper investigates and validates the impact of fine-tuning on large language model performance, focusing on parameter-efficient methods (LoRA and QLoRA) \cite{lora,intrinsic-dim,qlora}. We evaluate model capabilities across three key domains: (1) commonsense reasoning (HellaSwag \cite{hellaswag}), (2) mathematical reasoning (GSM8K\footnotemark[2] \cite{gsm8k}), and (3) multi-domain knowledge (MMLU-CS\footnotemark[3] \cite{mmlu}). 

Our findings demonstrate that: (1) LoRA-based methods effectively improve task-specific performance while maintaining computational efficiency, and (2) performance strongly depends on alignment between fine-tuning dataset and benchmark tasks. The study provides both theoretical insights into parameter-efficient mechanisms \cite{qlora} and practical guidance for developers implementing efficient LLM adaptation with limited resources.

 \end{abstract}
\footnotetext[1]{HellaSwag: A dataset for measuring commonsense natural language inference with challenging sentence completion tasks.}
\footnotetext[2]{Grade School Math 8K (GSM8K): A dataset of 8.5K high-quality grade school math word problems requiring multi-step reasoning.}
\footnotetext[3]{MMLU (Massive Multitask Language Understanding): A benchmark testing knowledge across 57 subjects. CS refers to the Computer Science subset.}
\footnotetext[4]{Source: default precision of Stata output number, see \url{https://www.stata.com/manuals13/ddatatypes.pdf}}

\renewcommand{\Keywords}{{\small \emph{Keywords}: \lmjkeywords}}

\Keywords

\section{Introduction}
Training large neural networks with hundreds of billions of parameters is computationally expensive and resource-intensive. As the size of models grows, so do the costs associated with training them. Parameter-efficient fine-tuning techniques, such as LoRA \cite{lora,intrinsic-dim} and QLoRA \cite{qlora}, have emerged as promising solutions to reduce these costs while maintaining or even improving model performance.

\subsection{Research Questions}
While this paper does not investigate the inner workings of LoRA, we leverage its core principle to examine whether updating only a small proportion of parameters in large language models can enhance computational efficiency without sacrificing performance.

\begin{itemize}
    \item Does fine-tuning lead to significant improvements in the performance of large language models? If so, what type of task it improved? And how the fine-tune process can be designed? (i.e. Is the type of the fine-tunning dataset used on the fine-tuning matters? Will the fine-tuning give promise to retaining the other LLMs' abilities like "Commonsense reasoning" or "Mathematical reasoning"? )
\end{itemize}

By answering these questions, this work aims to empower individual developers and researchers to effectively fine-tune smaller-scale language models using limited datasets.

\subsection{Objectives \& Literature review}
The primary objectives of this study are:
\begin{itemize}
    \item To explain the mathematical/engineering intuition of QLoRA (Quantization + LoRA) and how this contribute to efficient fine-tuning for LLM.
    
    \item Design an experiment and evaluate the impact of parameter-efficient fine-tuning on model performance across commonsense reasoning (HellaSwag), mathematical reasoning (GSM8K), and domain knowledge (MMLU) benchmarks
\end{itemize}

\subsubsection{Key Concepts}
Before we understand what is Large language model fine-tune, we have to get familiar with the concepts like: "Precision" \& "Quantization" because they are important technique to reduce the size of the large language model and also run faster in the GPU (take less time to load the data).
\begin{itemize}
    \item \textbf{Precision}: Refers to the bit-width of numerical representations (e.g., 32-bit floats). Higher precision preserves more information but increases memory usage and computation time \cite{8bit}. Recent researchers have overcome the bottleneck of the memory loading for large language model, and developed various optimizers that use 8-bit statistics while maintaining the performance levels of using 32-bit optimizer states \cite{8bit-optimizer}. As we all know, the different types of neural network can be represented as different architectures that have the weight as the elementary unit. And these weight are floating points numbers (i.e. float32 in Python). Consider the software Stata, the default setup for an output number is 7 digits of accuracy\footnotemark[4]. The longer output digits it provide, the higher data type size/computer memory is required. 
    
    \item \textbf{Quantization}: The process of reducing numerical precision (e.g., from 32-bit to 4-bit) to decrease model size and accelerate inference while minimizing accuracy loss \cite{4bit}. Model size follows:
    \begin{equation}
    \label{eq:model_size}
    \text{Model size} = \text{size}_{\text{datatype}} \times \text{num weights}
    \end{equation}

    As shown in Equation~\ref{eq:model_size}, the model size depends on the data type size and the number of weights. So in this case, people has come up with idea that using quantization technique which only need less memory with lowerer precision, but still maintain the model performance. In general, it is a technique that used to make the Model size smaller by lowering the precision without sacrificing the model performance \cite{8bit}. The takeaway here is that when loading the large language model, use the 4-bit quantization is almost universally optimal \cite{4bit}.
    
    \item \textbf{Low-Rank Adaptation (LoRA)}: Taking advantage of open-source pre-trained language models (PLMs) for the downstream similar task (i.e. the BERT is a base model, and fine-tuning the BERT model means making it suitable for certain specific task) is now the prevalent paradigm in natural language processing. The most common way to adapt general-purpose PLMs to downstream tasks is to fine-tune all the model parameters (full fine-tuning).\cite{unified-view} However, recent work has proposed a variety of parameter-efficient transfer learning methods that only fine-tune the necessary parameters to attain strong performance. There are generally 4 types of fine-tuning: (1)Transfer learning (Fine-tune last layer for downstream), (2)Adapter Layer (Adding new intermediate modules within the layers of pre-trained model), (3)Prefix Fine-tuning(Prompt Engineering), (4)LoRA(Rank decomposition). These different methods will be discussed in the Methods section in more details. \cite{adapter,prefix,attention,lora}

    Considering the computation efficiency; we want to reduce the number of computing-intensive parameters as many as possible. There is a basic example for the LoRA (one of the parameter-efficient methods):
Suppose you have a pre-trained language model with a weight matrix \( W \) of size \( 1000 \times 1000 \). Fine-tuning this model directly would require updating all 1,000,000 parameters in \( W \), which is computationally expensive.

\textbf{LoRA Approach}

Instead of updating all parameters in \( W \), LoRA introduces two low-rank matrices \( A \) and \( B \):
\begin{itemize}
    \item Let \( A \) be of size \( 1000 \times 10 \) (rank \( r = 10 \)).
    \item Let \( B \) be of size \( 10 \times 1000 \) (rank \( r = 10 \)).
\end{itemize}

Now, instead of updating 1,000,000 parameters, you only need to update:
\begin{equation}
\text{Number of trainable parameters} = (1000 \times 10) + (10 \times 1000) = 20,000
\end{equation}
This is a significant reduction in the number of trainable parameters.
\end{itemize}






\subsubsection{QLoRA Mathematical Formulation}
Given a pretrained weight matrix $W \in \mathbb{R}^{d \times d}$, LoRA learns:
\begin{itemize}
    \item $A \in \mathbb{R}^{d \times r}$ and $B \in \mathbb{R}^{r \times d}$ where $r \ll d$ (typically $r=8$)
    \item The update becomes $\Delta W = AB$ instead of full $d^2$ parameters
    \item Forward pass computes:
    \begin{equation}
    y = (W + AB)x
    \end{equation}
\end{itemize}

QLoRA extends this by quantizing $W$ to 4-bit precision \cite{qlora}, further reducing memory requirements while maintaining the low-rank adaptation benefits.

\section{Methods}

\subsection{Related work for large Model Training and Fine-Tuning}

There are several different fine-tunning techniques:

\textbf{Transfer Learning (Fine-tune last layer for downstream)}: The traditional way of transfer learning was to simply freeze all parameters and weights within the pre-trained model, and then add a new component at the right hand side. So in this case, the output of the pre-trained model will become the input of the new component(could be a layer or new neural network architecture just defined for a specific downstream task). The downside for this method is that our downstream model can only access the "output embedding" of the pre-train model and can not adjust the weight of the pre-trained model when doing backpropagation. \cite{transfer}

\textbf{Adapter Layers (Add new intermediate modules within the layers of pre-trained mdoel)}: Google research paper has proposed that the fine-tuned can be conducted within the original pre-trained model, which new modules can be inserted between the layers. However, the drawback is that this new added layer would increase the latency during the inference of the model and general computational efficience become lower. \cite{adapter}

\textbf{Prefix Tuning (Prompt Engineering)}: Another fine-tune solution is to modify the prefix of every layer of the pre-trained model, this is similar to do a feature engineering which simply optimizes the input of each layer. \cite{prefix}

\textbf{LoRA (Rank decomposition)}: This is the most commonly used method. It perform a rank decomposition on the updated weight metrics. With this optimization direction, this method actually really suitable for the transformer related architecture. Those attention mechanism always require the weight matrix to compute the embedding. And rank decomposition will always be applied to these updated weighted matrix. \cite{lora,attention} 

And in this paper, we will focus on using the most common one: QLoRA, which is the combination of quantization and "Low-rank adaptation".

The full name of LoRA is "Low-rank adaptation", which is not a new concept, and it has been widely used in many traditional machine learning areas: "sparse-matrix problem" in the recommendation area, "Sense compressing" in the image compression. Facebook research team \cite{intrinsic-dim} provided an example of how to apply this technique in the large language model for fine-tuning purpose. For better understanding, "rank" is a mathematical concept that represents the minimum number of independent rows/columns. This is an important mathematical characteristic when manipulating the matrix from various domains' applications. 

Basically, the engineer wants to use a smaller matrix to approximate the undetermined matrix and hence reducing computing complexity.

And this paper is not studying how the LoRA works, but by leveraging the idea of this method, we want to see if this is actually help to "Slimming Down" the Large language model while improving the computing efficiency of the sparse matrix; downstream tasks do not need to tune all the parameters within the whole parameter space for that specific model. Instead, it can be transformed into a much smaller set of weights to achieve a good performance model. They find that the larger the model, the lower the intrinsic dimension \cite{intrinsic-dim}.

In traditional fine-tuning, the weight matrix \( W \) of a layer is updated directly:
\[
W_{\text{new}} = W_{\text{old}} + \Delta W
\]
where \( \Delta W \) is the full-rank weight update.

LoRA approximates the weight update \( \Delta W \) using two low-rank matrices \( A \) and \( B \):
\[
\Delta W = A \cdot B
\]
where:
\begin{itemize}
    \item \( A \in \mathbb{R}^{d \times r} \) is a low-rank matrix with rank \( r \).
    \item \( B \in \mathbb{R}^{r \times d} \) is another low-rank matrix with rank \( r \).
    \item \( r \ll d \), meaning the rank \( r \) is much smaller than the original dimension \( d \).
\end{itemize}

\subsubsection{Final Weight Update with LoRA}

The updated weight matrix \( W_{\text{new}} \) is computed as:
\[
W_{\text{new}} = W_{\text{old}} + A \cdot B
\]
Here, only \( A \) and \( B \) are trainable, while \( W_{\text{old}} \) remains frozen.

\subsubsection{Forward Pass with LoRA}
During the forward pass, the output \( y \) is computed as:
\[
y = (W_{\text{old}} + A \cdot B) \cdot x
\]
where \( x \) is the input to the layer. And combining the quantization advantage, this is how we utilize the QLoRA to fine tune our large language model. \cite{qlora}

\subsection{Transformer Architecture \& Parameter-Efficient Tuning} 
\label{sec:arch}

\begin{figure}[h!]
\centering

\includegraphics[width=0.9\textwidth]{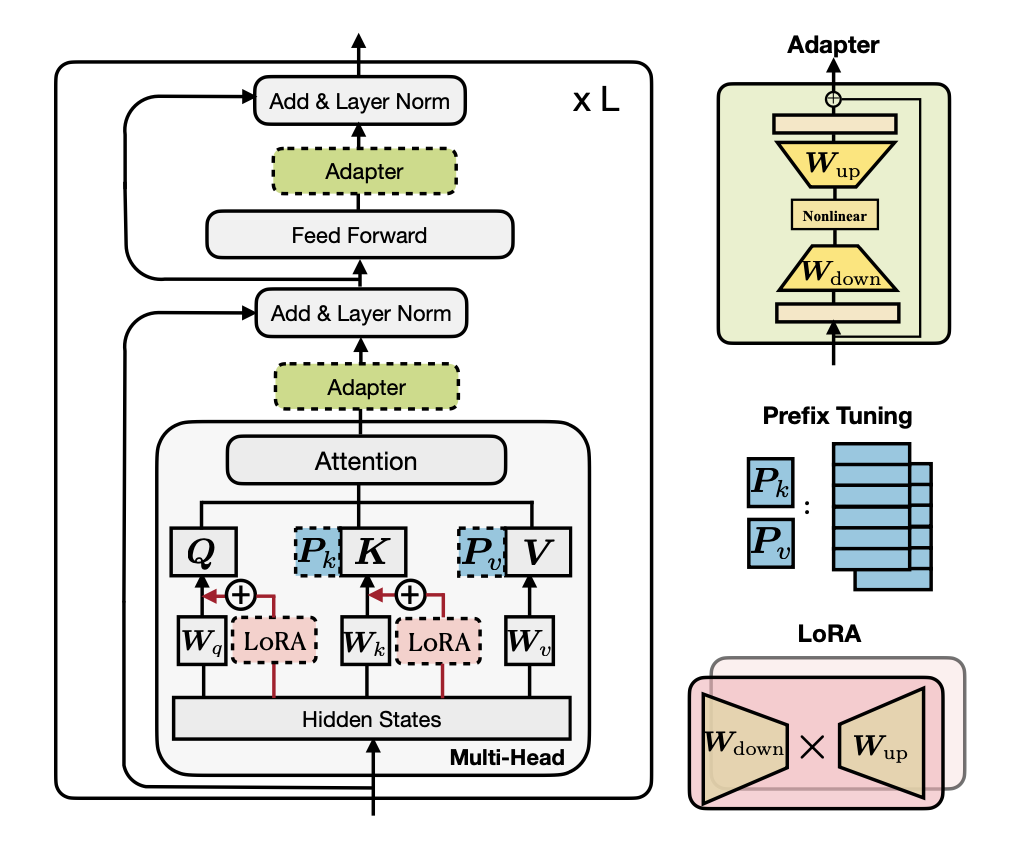}

\caption{Transformer architecture with parameter-efficient tuning methods. Dashed blocks represent added modules: (1) LoRA updates weight matrices through low-rank decomposition, (2) Adapters insert projection layers, (3) Prefix Tuning prepends learnable vectors to attention keys/values. Picture Source from \cite{unified-view}}
\label{fig:peft-methods}
\end{figure}

Modern transformer-based LLMs process inputs through stacked layers containing two core components:

\subsubsection{Recap for Transformer Base Architecture}
\begin{itemize}
\item \textbf{Multi-Head Attention \cite{attention,bert}}: Computes contextual relationships via:
\[
\text{Attention}(Q,K,V) = \text{softmax}\left(\frac{QK^T}{\sqrt{d_k}}\right)V
\]
where queries (Q), keys (K), and values (V) are projections of hidden states.

\item \textbf{Feed-Forward Network (FFN)}: Applies nonlinear transformations:
\[
\text{FFN}(x) = \text{GeLU}(W_2 \cdot \text{GeLU}(W_1x + b_1)) + b_2
\]

\item \textbf{Residual Connections}: Stabilize training through:
\[
x_{out} = \text{LayerNorm}(x + \text{Sublayer}(x))
\]
\end{itemize}

\subsubsection{Efficient Tuning Mechanisms} 
There are three parameter-efficient strategies (Fig.~\ref{fig:peft-methods}), in this paper, we only apply the QLoRA (quantization version of Low-Rank Adaptation):

1. \textbf{LoRA (Low-Rank Adaptation)}:
\begin{itemize}
\item Decomposes weight updates: $\Delta W = AB$ where $A \in \mathbb{R}^{d \times r}$, $B \in \mathbb{R}^{r \times d}$ 
\item Applied to Q/K/V projections in attention heads
\item Trains only $A$ and $B$ (the trainable \% of total parameters depend on the hyper parameter "rank" )
\end{itemize}

2. \textbf{Adapters}:
\begin{itemize}
\item Inserts bottleneck layers after attention/FFN:
\[
x_{adapted} = x + W_{up} \cdot \text{GeLU}(W_{down} \cdot x)
\]
where $W_{down} \in \mathbb{R}^{d \times r}$, $W_{up} \in \mathbb{R}^{r \times d}$
\end{itemize}

3. \textbf{Prefix Tuning}:
\begin{itemize}
\item Prepends learnable vectors $P_k$, $P_v$ to attention keys/values:
\[
K' = [P_k; K], \quad V' = [P_v; V]
\]
\end{itemize}

\subsubsection{Merge Process} 
Post-training, adapter parameters are integrated into base weights through:
\[
W_{\text{merged}} = W_{\text{base}} + \Delta W_{\text{adapter}}
\]
preserving the original model size (1.2B parameters) while encoding new capabilities. This enables deployment without additional computational overhead.

\subsection{Foundation Model \& Training Data for fine-tune purpose}

In this study, we compare the performance of two base models: \textbf{Llama 3.2 1B} and \textbf{TinyLlama 1.1B}. Both models are pre-trained and fine-tuned using the \textbf{Alpaca dataset} \footnotemark[6] (high-quality instruction-following data), as summarized in Table~\ref{tab:experiment_setup}. These models are downloaed from Hugging Face \footnotemark[5] and we have been granted access to these models through acknowledgment. 

\subsubsection{Model Selection Justification}

Llama 3.2 \footnotemark[9] provides a stronger baseline (more advanced model architecture), while TinyLlama tests efficiency limits (less advanced model).

We compare \textbf{Llama 3.2 1B} (1.24B params) and \textbf{TinyLlama 1.1 1B} because:
\begin{itemize}
    \item \textbf{Controlled Scaling}: Similar parameter counts (1.24B vs 1.1B) isolate architectural effects from pure size differences
    \item \textbf{Practical Accessibility}: Both fit consumer-grade hardware (T4 GPU), validating our methods' real-world applicability
    \item \textbf{Architectural Spectrum}: Contrasts Llama 3.2's optimized attention mechanisms with TinyLlama's simplified design, testing whether the QLoRA fine-tune can improve the ability of the less advanced model (out-perform the more advanced one (without fine-tuning))
    \item \textbf{Robustness of the QLoRA method}: Contrasting architectures highlight the robustness of the QLoRA finetune method.
\end{itemize}

\textbf{Llama 3.2 1B} serves as our performance ceiling - a modern architecture where we test capability retention during sparse updates. \textbf{TinyLlama 1.1B} represents the efficiency frontier, where we measure if parameter reduction enables competitive performance through focused adaptation on fine-tune dataset.

Plus,

\textbf{Llama 3.2 1B \footnotemark[5]} is a 1.24B params billion parameter model from the Llama family, known for its balance between performance and computational efficiency. It serves as a baseline for comparison in this study.

\textbf{Llama 1.1 1B} is a smaller variant of the Llama family with 1.1 billion parameters, designed for resource-constrained environments. Despite its smaller size, it is fine-tuned on a larger subset of the Alpaca dataset. (50,000 samples)

\footnotetext[5]{meta-llama/Llama-3.2-1B, Hugging Face. \url{https://huggingface.co/meta-llama/Llama-3.2-1B}}

\footnotetext[6]{Alpaca: A Strong, Replicable Instruction-Following Model, Stanford. \url{https://crfm.stanford.edu/2023/03/13/alpaca.html}}

\footnotetext[8]{Finetune framework for LLM, Unsloth. \url{https://github.com/unslothai/unsloth}}

\footnotetext[9]{Llama 3.2: Revolutionizing edge AI and vision with open, customizable models, Meta. \url{https://ai.meta.com/blog/llama-3-2-connect-2024-vision-edge-mobile-devices/}}

All experiments were conducted using the following hardware and software setup:
\begin{itemize}
    \item \textbf{Hardware}: Google Colab with Tesla T4 GPU acceleration.
    \item \textbf{Software}: Python 3 environment.
    \item \textbf{Fine-Tuning Framework}: QLoRA using the Sloth framework \footnotemark[8].
\end{itemize}


\begin{table}[h!]
\centering
\captionsetup{justification=centering} 
\caption{Experiment Setup: Foundation Models, Training Data, and Fine-Tuning Methods}
\label{tab:experiment_setup}
\begin{tabular}{lllll}
\toprule
\textbf{Model} & \textbf{Model Size} & \textbf{Training Data} & \textbf{Training Size} & \textbf{Method} \\
\midrule
\textbf{Base Llama 3.2 1B} & 1.24B parameters & None (Pre-trained) & - & - \\
\textbf{Fine-tuned Llama 3.2 1B} & 1.24B parameters & Alpaca Dataset & 2,000 samples & QLoRA \\
\textbf{TinyLlama 1.1 1B} & 1.1B parameters & Alpaca Dataset & 50,000 samples & QLoRA \\
\bottomrule
\multicolumn{5}{l}{* limited to 2,000 due to hardware constraints, TinnyLlama 1.1B takes around 14 hours to finetune} \\
\end{tabular}
\end{table}

By doing this experiment, we can determine:

(1) after fine-tune, whether TinyLlama 1.1B with less advanced architecture can outperform the Llama 3.2 1B (advanced version for Llama family);

(2) after fine-tune, whether the Llama 3.2 1B can outperform it base version (without fine-tuning);




    



\subsection{Benchmark Datasets and Evaluation Metrics}

We employ three established benchmarks with full operationalization of all metrics and variables, the evaluation metrics are calculated by leveraging the GitHub open-source framework "lm-evaluation-harness" \cite{eval-harness} \footnotemark[7] :

\begin{itemize}
    \item \textbf{HellaSwag Dataset \cite{hellaswag} (Commonsense Reasoning)}
    \begin{itemize}
        \item \textbf{Purpose}: Evaluate preservation of everyday reasoning
        \item \textbf{Metric Variables}:
        \begin{itemize}
            \item \( p_i \): Model's predicted continuation for \( i^{th} \) sample
            \item \( t_i \): Human-preferred ground truth continuation
            \item \( n \): Total samples (\( n = 10,000 \))
            \item \( f(x) \): Normalization function = lowercase(remove\_punctuation(x))
            \item \( A^{\text{Base}}_{\text{norm}} \): Normalized accuracy of base model
            \item \( A^{\text{FT}}_{\text{norm}} \): Normalized accuracy after fine-tuning
        \end{itemize}
        \item \textbf{Metrics}:
        \[
        A_{\text{norm}} = \frac{\sum_{i=1}^n \mathbb{I}(f(p_i){=}f(t_i))}{n} \times 100\% 
        \]
        \[
        \Delta A = A^{\text{FT}}_{\text{norm}} - A^{\text{Base}}_{\text{norm}} 
        \]
    \end{itemize}
    
    \item \textbf{GSM8K Dataset \cite{gsm8k} (Mathematical Reasoning)}
    \begin{itemize}
        \item \textbf{Purpose}: Quantify mathematical capability retention
        \item \textbf{Metric Variables}:
        \begin{itemize}
            \item \( p_i \): Numerical answer extracted from model output
            \item \( t_i \): Ground truth numerical value
            \item \( \epsilon \): Tolerance threshold (\( \epsilon = 0.01 \))
            \item \( A^{\text{Base}}_{\text{flex}} \): Base model's flexible accuracy
            \item \( A^{\text{FT}}_{\text{flex}} \): Fine-tuned model's flexible accuracy
        \end{itemize}
        \item \textbf{Metrics}:
        \[
        A_{\text{strict}} = \frac{\sum \mathbb{I}(p_i{=}t_i)}{n} \times 100\% \quad \text{(Exact Match)}
        \]
        \[
        A_{\text{flex}} = \frac{\sum \mathbb{I}(|p_i{-}t_i|{<}\epsilon)}{n} \times 100\% \quad \text{(Flexible Extract)}
        \]
        \[
        \text{FR} = \left(1 - \frac{A^{\text{FT}}_{\text{flex}}}{A^{\text{Base}}_{\text{flex}}}\right) \times 100\% \quad \text{(Forgetting Rate)}
        \]
    \end{itemize}
    
    \item \textbf{MMLU Dataset \cite{mmlu} (Domain Knowledge)}
    \begin{itemize}
        \item \textbf{Purpose}: Measure factual knowledge retention
        \item \textbf{Metric Variables}:
        \begin{itemize}
            \item \( p_i \): Model's selected option (A/B/C/D)
            \item \( t_i \): Correct option identifier
            \item \( z \): Z-score (\( z = 1.96 \))
            \item \( A^{\text{Base}} \): Base model accuracy
            \item \( A^{\text{FT}} \): Fine-tuned model accuracy
            \item \( \text{CI}_{\text{Base}} \): Base model confidence interval
            \item \( \text{CI}_{\text{FT}} \): Fine-tuned model confidence interval
        \end{itemize}
        \item \textbf{Metrics}:
        \[
        A = \frac{\sum \mathbb{I}(p_i{=}t_i)}{n} \times 100\% \quad \text{(Accuracy)}
        \]
        \[
        \Delta K = A^{\text{Base}} - A^{\text{FT}} \pm \sqrt{\text{CI}_{\text{Base}}^2 + \text{CI}_{\text{FT}}^2} \quad \text{(Knowledge Loss)}
        \]
    \end{itemize}
\end{itemize}

\footnotetext[7]{Implementation Details: 
\begin{itemize}
\item All metrics computed using lm-evaluation-harness \cite{eval-harness}
\item Statistical significance: \( p < 0.05 \) via paired t-test
\item Confidence intervals: 95\% level (\( \alpha = 0.05 \))
\end{itemize}
}

\section{Results}

As shown in Table~\ref{tab:performance_comparison}, parameter-efficient fine-tuning produces divergent outcomes across capability domains, quantified through our formal metrics:

\begin{table}[h!]
\centering
\caption{Performance Comparison of Base and Fine-Tuned Models}
\label{tab:performance_comparison}
\begin{tabular}{llccc}
\hline
\textbf{Task} & \textbf{Metric} & \textbf{Base Llama 3.2 1B} & \textbf{Fine-tuned Llama 3.2 1B (2k)} & \textbf{TinyLlama 1.1B (50k)} \\
\hline
\multirow{3}{*}{GSM8K \footnotemark[2]} 
 & Flexible Accuracy & \textbf{33.51\% $\pm$ 1.30\%} & 3.71\% $\pm$ 0.52\% & 2.81\% $\pm$ 0.45\% \\
 & Strict Accuracy & \textbf{33.51\% $\pm$ 1.30\%} & 3.11\% $\pm$ 0.48\% & 2.20\% $\pm$ 0.40\% \\
 & Forgetting Rate (FR) & - & 88.92\%* & - \\
\hline
\multirow{3}{*}{HellaSwag \footnotemark[1]} 
 & Accuracy & 45.23\% $\pm$ 0.50\% & 45.61\% $\pm$ 0.50\% & \textbf{45.84\% $\pm$ 0.50\%} \\
 & Normalized Accuracy & 60.77\% $\pm$ 0.49\% & \textbf{61.20\% $\pm$ 0.49\%} & 59.26\% $\pm$ 0.49\% \\
 & Ability Augmentation ($\Delta A$) & - & +0.43\% & - \\
\hline
\multirow{2}{*}{MMLU CS \footnotemark[3]} 
 & Accuracy & \textbf{47.00\% $\pm$ 5.02\%} & 34.00\% $\pm$ 4.76\% & 27.00\% $\pm$ 4.46\% \\
 & Knowledge Loss ($\Delta K$) & - & 13.00\% $\pm$ 6.93\%* & - \\
\hline
\end{tabular}
\end{table}

\footnotetext[1]{Ability Augmentation calculated as $\Delta A = A^{\text{FT}}_{\text{norm}} - A^{\text{Base}}_{\text{norm}}$}
\footnotetext[2]{Forgetting Rate (FR) = $\left(1 - \frac{A^{\text{FT}}_{\text{flex}}}{A^{\text{Base}}_{\text{flex}}}\right) \times 100\%$}
\footnotetext[3]{Knowledge Loss $\Delta K = A^{\text{Base}} - A^{\text{FT}} \pm \sqrt{\text{CI}_{\text{Base}}^2 + \text{CI}_{\text{FT}}^2}$}
\footnotetext[4]{- indicates metric not applicable/comparable due to architectural differences}

\textbf{1. Catastrophic Forgetting in Mathematical Reasoning}
\begin{itemize}
\item \textit{Key Metric}: Forgetting Rate (FR) = 88.92\%* (GSM8K Flexible Extract)
\item \textit{Base vs. Fine-tuned}: 33.51\% $\pm$ 1.30\% → 3.71\% $\pm$ 0.52\%
\item \textit{Interpretation}: The FR metric reveals near-total loss of mathematical reasoning (89\% relative decrease), exceeding typical catastrophic forgetting thresholds (more than 50\% FR)
\end{itemize}

\textbf{2. Preserved Commonsense Reasoning Ability}
\begin{itemize}
\item \textit{Key Metric}: Ability Augmentation ($\Delta A$) = +0.43\% (HellaSwag Normalized Accuracy)
\item \textit{Base vs. Fine-tuned}: 60.77\% $\pm$ 0.49\% → 61.20\% $\pm$ 0.49\%
\item \textit{Interpretation}: Stable $\Delta A$ suggests structural knowledge persistence despite:
\begin{itemize}
\item Architectural constraints prioritizing conversational ability
\item Shared representational space for commonsense patterns
\end{itemize}
\end{itemize}

\textbf{3. Domain Knowledge Degradation}
\begin{itemize}
\item \textit{Key Metric}: Knowledge Loss ($\Delta K$) = 13.00\% $\pm$6.93\%* (MMLU CS)
\item \textit{Base vs. Fine-tuned}: 47.00\% $\pm$5.02\% → 34.00\% $\pm$4.76\%
\item \textit{Interpretation}: Significant but partial unlearning (28\% relative loss) indicates:
\begin{itemize}
\item Factual knowledge's moderate robustness to parameter updates
\item Residual semantic connections despite objective misalignment
\end{itemize}
\end{itemize}

\textbf{4. Architectural Efficiency Analysis}
\begin{itemize}
\item \textit{Cross-Model Comparison}: Despite 25× less data, fine-tuned Llama 3.2 (34.00\% $\pm$4.76\%) outperforms TinyLlama (27.00\% $\pm$4.46\%) on MMLU CS
\item \textit{Key Insight}: The accuracy gap (7 percentage point ) persists with:
\begin{itemize}
\item TinyLlama's simplified architecture (less complex model architecture)
\item Substantially larger fine-tuning dataset (50k vs 2k samples)
\end{itemize}
\end{itemize}

\section{Discussion}

Fine-tuning strategies must balance new capabilities with knowledge preservation, especially for smaller models. Future research should explore mixed fine-tuning datasets, parameter-efficient techniques, and targeted evaluation. Research reveals critical trade-offs that must be considered when fine-tuning LLMs:

Task-specific capabilities: Fine-tuning without task-specific examples can completely erase specialized capabilities (as seen with Table~\ref{tab:performance_comparison} - "Base Llama 3.2 1B" and "Fine-tuned Llama 3.2 1B (2k)" before and after mathematical reasoning, Forgetting Rate (FR) = *88.92\% )

Representational competition: Neural resources are finite - improving one capability often comes at the cost of others

Data efficiency vs. knowledge retention: More data doesn't necessarily overcome architectural limitations

 \bibliography{x}

\end{document}